\documentclass[10pt,twocolumn,letterpaper]{article}
\usepackage[accsupp]{axessibility} 
\usepackage{iccv}
\usepackage{times}
\usepackage{epsfig}
\usepackage{graphicx}
\usepackage{amsmath}
\usepackage{amssymb}
\usepackage{multirow}
\usepackage{subcaption}
\usepackage[pagebackref=true,breaklinks=true,letterpaper=true,colorlinks,bookmarks=false]{hyperref}

\iccvfinalcopy % *** Uncomment this line for the final submission

\def\iccvPaperID{10209} % *** Enter the ICCV Paper ID here

\ificcvfinal\fi

\begin{document}

%%%%%%%%% TITLE
\title{Flexible Visual Recognition by Evidential Modeling of \\Confusion and Ignorance}

\author{
Lei Fan\textsuperscript{1},
Bo Liu\textsuperscript{2}, 
Haoxiang Li\textsuperscript{2},
Ying Wu\textsuperscript{1} and Gang Hua\textsuperscript{2}\\
\textsuperscript{1}Northwestern University \textsuperscript{2}Wormpex AI Research\\
{\tt\small leifan@u.northwestern.edu,yingwu@northwestern.edu,\{richardboliu,lhxustcer,ganghua\}@gmail.com}\\
}

\maketitle
\ificcvfinal\thispagestyle{empty}\fi

%%%%%%%%% ABSTRACT
\begin{abstract}
In real-world scenarios, typical visual recognition systems could fail under two major causes, i.e., the misclassification between known classes and the excusable misbehavior on unknown-class images.
To tackle these deficiencies, flexible visual recognition should dynamically predict multiple classes when they are unconfident between choices and reject making predictions when the input is entirely out of the training distribution.
Two challenges emerge along with this novel task. 
First, prediction uncertainty should be separately quantified as confusion depicting inter-class uncertainties and ignorance identifying out-of-distribution samples.
Second, both confusion and ignorance should be comparable between samples to enable effective decision-making.
In this paper, we propose to model these two sources of uncertainty explicitly with the theory of Subjective Logic. Regarding recognition as an evidence-collecting process, confusion is then defined as conflicting evidence, while ignorance is the absence of evidence.
By predicting Dirichlet concentration parameters for singletons, comprehensive subjective opinions, including confusion and ignorance, could be achieved via further evidence combinations.
Through a series of experiments on synthetic data analysis, visual recognition, and open-set detection, we demonstrate the effectiveness of our methods in quantifying two sources of uncertainties and dealing with flexible recognition.
\end{abstract}

%%%%%%%%% BODY TEXT
\section{Introduction}
\label{sec:intro}

\begin{figure}[t]
\centering
\includegraphics[width=1\linewidth]{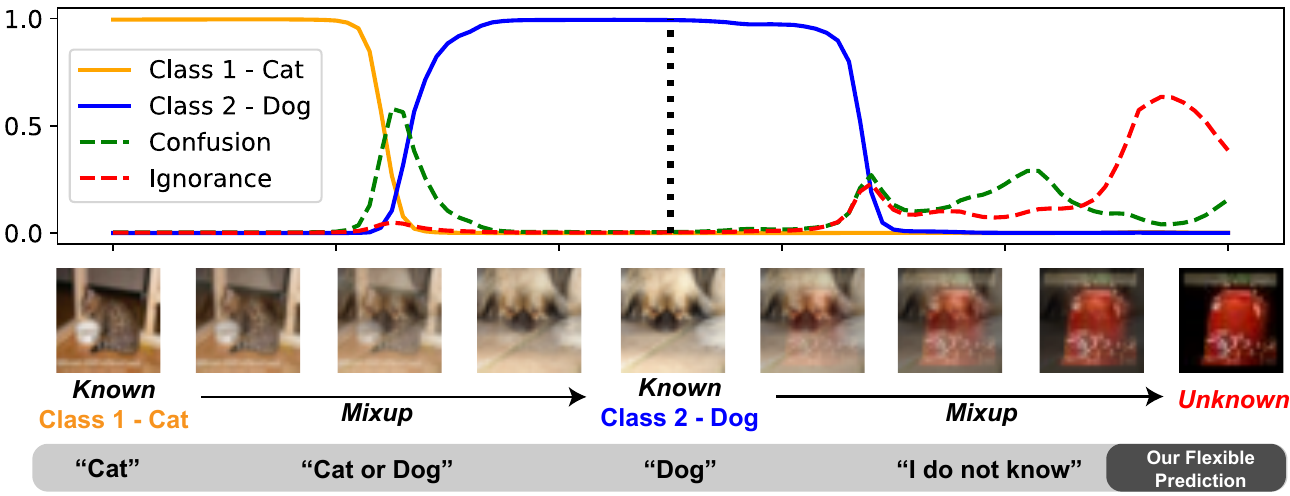}
\caption{Classification of the proposed approach on images interpolated from a {\it{known-known-unknown}} triplet. Ignorance reflects the lack of evidence, whereas confusion is caused by conflicting evidence, \ie, evidence that fails to provide discrimination between specific classes.
A flexible visual recognition system could provide combined predictions when having large confusion and reject making predictions for unknown-class samples.
Note the mixup images are for illustrative purposes and are not a requisite in our training.}
\label{fig:firstsight_a}
\vspace{-2mm}
\end{figure}

% Paragraph 1 - Intro to why uncertainty estimation matters and why we need flexible visual recognition.
% 1. A typical recognition system provides one single prediction. However, in an open-world scenario, it is critical to obtain reliable uncertainty estimates for downstream tasks, including...
% 2. A robust and flexible recognition system should ... Give an example when measuring traditional uncertainty is not enough and confusion could help.

When employing visual classifiers in open-world conditions, obtaining reliable uncertainty estimations could significantly benefit downstream tasks, including autonomous driving~\cite{amini2020deep,djuric2020uncertainty,michelmore2018evaluating}, medical diagnosis~\cite{mehrtash2020confidence,wang2019aleatoric}, and embodied intelligence~\cite{patro2019u,malinowski2014multi}. Recent uncertainty quantification techniques~\cite{zhou2017evidence,gal2015bayesian,lakshminarayanan2017simple,liu2020simple,malinin2019reverse,malinin2018predictive,charpentier2020posterior,sensoy2020uncertainty} have achieved notable progress toward this goal by saying {\it{``I do not know''}} when the testing distributions differ from the training. However, besides directly giving no prediction, a more flexible and informative visual recognition system could also give combined predictions when possible, implying the correct answer is one of its predictions but uncertain. Naturally, the capability of rejecting or providing unspecific predictions demands separately measuring different sources of uncertainties, \ie, ignorance and confusion, if seen from the Subjective Logic~\cite{josang2016subjective} perspective. Furthermore, to enable flexible recognition, both uncertainties should possess the virtue of comparability between samples and in-sample additivity.

% Paragraph 2 - Confusions and ignorance
In evidential deep learning, the training of a recognition model could be regarded as an evidence-collecting process~\cite{josang2016subjective,sensoy2018evidential}. Unlike ignorance describing a total lack of evidence, confusion is defined as conflicting evidence, which mandates the existence of multiple hypotheses in the frame of discernment. In other words, we cannot assess confusion for a single-class classification problem. The mass of confusion between two classes then reflects shared features that contribute to both classes while not discriminative. Likewise, confusion exists for all combinations of classes larger than two. Unlike typical visual classifiers that only predict a single output, flexible predictions could be obtained if we could combine singleton belief derived from class-exclusive evidence with their inter-class confusion.

% Paragraph 3 - Recent approaches on uncertainty estimation
% 1. Classic uncertainty estimation categories.
% 2. Evidential uncertainty estimation models.
% 3. For EDL, confusion and ignorance are estimated with a single term.
With great potential for explicitly estimating confusion and ignorance, this area is still under-explored for deep visual classifiers. Recent methods regard uncertainties as the degree of mismatch between training and testing distributions, which comprise but do not distinguish between confusion and ignorance. Deep Bayesian models, including dropout~\cite{gal2015bayesian,kendall2015bayesian} and ensemble-based approximations~\cite{lakshminarayanan2017simple,wenzel2020hyperparameter,ashukha2020pitfalls}, require multiple forwards to estimate the posterior predictive distribution. Evidential models~\cite{sensoy2018evidential,amini2020deep,corbiere2021beyond,bao2021evidential} predict parameters of the posterior of class distribution directly. However, these models regard uncertainty as a whole term covering both confusion and ignorance, making it infeasible to perform flexible visual recognition further. 

% Paragraph 4 - The proposed method.
% 1. What are the challenges of modeling confusion? 
%   c. Too many terms for confusion. And how do we handle it? By adopting the rule of evidence combination, we formulate the problem under a k-dimensional space for a k-class classification problem. k plausibility functions under the same frame.
% 2. Figure 1 Figure 2.
% 3. As we only have a single deterministic label, how do we form opinions? Generate singleton beliefs from plausibility functions and then use a neural network to learn conjugated Dirichlet prior parameters.
% In Fig.~\ref{fig:firstsight_a}, a confusion surge is expected when the input 

Distinct from existing uncertainty quantification methods, the proposed method models confusion and ignorance for each sample separately, which provides valuable information to facilitate various visual tasks, including flexible visual recognition. An illustrative example with the prediction of our method is shown in Fig.~\ref{fig:firstsight_a}. Under the theory of Subjective Logic~\cite{barnett2008computational,josang2016subjective}, confusion is defined as the shared evidence contributing to multiple categories while not discriminative between them, while ignorance is completely missing evidence. 

% As confusion happens among all class combinations, to avoid combinatorial complexity, plausibility functions are introduced to predict per-class evidence for each image. The confusion could be directly inferred through evidence combinations. Fig.~\ref{fig:firstsight_b} compares the behavior of our approach with the entropy of a standard network in the 2-dimensional space. As we can observe, the proposed method acts as a density estimator. Confusion happens between class boundaries, while ignorance is high for out-of-distribution data points. In addition, the proposed method does not require intricate adaptions to network architectures to achieve the desired properties.

% Paragraph 5 - Our contributions
% 1. As our method could achieve two different sources of uncertainties. Designing experiments is challenging due to the lack of ground truth. We transfer the problem.
% 2. Other experiments.
% 3. Our advantages.
The contribution of this paper could be summarized as follow: (1) The proposed method could explicitly predict two sources of uncertainties, \ie, confusion and ignorance, simultaneously for each sample. (2) The solution to confusion and ignorance is based on standard architectures, and the training does not rely on external information. (3) The effectiveness of the proposed method is extensively validated across different experiments, including studies on synthetic data, visual recognitions, and open-set detections.

%------------------------------------------------------------------------
\section{Related Work}
\label{sec:related}
\noindent {\bf{Uncertainty estimation.}}
Typical neural networks can not detect their own failure. However, this ability can be important in several real-world applications, like rejecting unseen samples, and providing prediction confidence, to name a few. Bayesian NN~\cite{gal2015bayesian,kwon2020uncertainty,tran2019bayesian,dusenberry2020efficient,kristiadi2020being} predicts epistemic uncertainty as the mutual information between model parameters and samples. By assuming a probabilistic prior on the network, it approximates prediction variance by sampling weight during inference. Several works~\cite{hora1996aleatory,wang2019aleatoric,kendall2017uncertainties} choose to model epistemic and aleatoric uncertainties separately. The Subjective Logic on which our method is established falls within the realm of epistemology instead of a frequentist (aleatoric) view. In other words, we focus on further separating epistemic uncertainty into confusion and ignorance.

Evidential deep learning~\cite{sensoy2018evidential,amini2020deep,bao2021evidential,corbiere2021beyond}, in contrast, proposes to learn the prior of the predictions directly. The prior, known as evidential prior, is interpreted as beliefs in Dempster-Shafer Threory~\cite{barnett2008computational}. In~\cite{corbiere2021beyond}, they model the first- and second-order uncertainties by introducing an auxiliary uncertainty network to approximate the difference between Dirichlet distributions. 

While uncertainty is provided to describe the variance of model prediction, it is not clear if uncertainty comes from different sources when dealing with in-distribution or out-of-distribution data. Orthogonal to previous approaches, we separate the uncertainty into confusion and ignorance in this work. Confusion depicts the uncertainty between different known classes, while ignorance decides whether the sample is unknown. With this separation, we can make dynamic predictions on known classes and reject unknown classes at the same time.
% 1. The motivation to quantify uncertainty, including possible applications.
% 2. Recent different works on this task.
%   a. Bayesian NN.
%   b. Deep ensemble methods (approximate BNN with Monte-Carlo methods).
%   c. Bayesian dropout (same as above).
%   d. Model and data uncertainty. Only introduce, do not emphasize.
%   e. Deep Gaussian process.
%   f. Prior network and posterior network (under Dirichlet distribution which is the conjugate prior of multinomial distribution).
%   g. Evidential deep learning methods (classification, regression, and its following works).
% 3. The need to separate different uncertainties (on open-set detection, on active learning, on hard-sample mining, etc).
%   a. First-order & second-order uncertainty.

\noindent {\bf{Open Set Recognition.}}
Machine learning models are usually designed with the closed-set assumption, where testing data shares the same distribution as the training. Open-set recognition (OSR)~\cite{bendale2016towards} introduces semantic shifts to the problem. Samples are from the classes that are not in the training set. Out-of-distribution (OOD)~\cite{hendrycks17baseline} detection introduces domain shifts to the testing set. In both of the settings, models should have the ability to reject unknown samples. 

In general, OSR and OOD methods reject unknown samples depending on reweighting outputs~\cite{bendale2016towards,ge2017generative,kong2021opengan,hendrycks17baseline,liang2018enhancing,wang2022vim}, getting better feature embedding metrics~\cite{liu2020few,chen2020learning,pmlr-v119-sastry20a,neal2018open}, and exploring reconstruction errors~\cite{oza2019c2ae,sun2020conditional,shao2020open,yoshihashi2019classification,perera2020generative}. These metrics are all related to the quality of classifier prediction. However, the recognition can fail on closed-set samples because of the existence of confusion. In this work, we show that when the confusion between known classes is adequately modeled, unknown samples can be identified more accurately.

\noindent {\bf{Conformal Prediction.}}
Parallel to our task, conformal prediction is a paradigm that could provide single or multiple predictions by empirically constructing confidence regions~\cite{angelopoulos2020uncertainty,angelopoulos2021gentle,vovk2005algorithmic}. 
% In~\cite{}, they propose a novel conformal prediction algorithm for selecting predictive sets by regularizing the scores of unlikely classes.  
However, conformal prediction is confined to closed scenarios without open samples, as the empirical quantile is established on a labeled validation set sampled from the same testing distribution.

% \subsection{Flexible Recognition}
% 1. What can we achieve with the proposed approach? Detect open-set and make indefinite predictions.
% 2. How do we handle open-set?
% 3. How do we make single or multiple predictions?
%   a. The meaning of making multiple predictions. Still the belief mass.
%   b. Greedily combine. It is optimal.
% Recall the two-fold motivations for doing flexible recognitions. The first is to reject samples lying entirely out of the training distribution. The second is making multiple predictions when necessary. This objective could be easily reached with the confusion and ignorance predicted by the proposed approach. An intuitive solution would be setting the belief threshold for outputs. The sample will be rejected if the ignorance is too large that no combination would exceed the threshold. And the model gives incrementally combined predictions if no singleton belief meets the bar.

%------------------------------------------------------------------------
\section{Flexible Recognition}
Flexible recognition aims to provide a classification model $\mathcal{M}$ that could deliver adaptive predictive sets. Specifically, the model rejects samples, \ie, making no prediction, when the input is entirely out of the training distribution, like an open-set sample. The model also should cautiously give a set of predictions with the true class contained in it when being unsure.

For a $K$-class classification problem, we formalize the flexible recognition system $\mathcal{M}(\cdot)$ as $\{y_1,\dots, y_k\}= \mathcal{M}(\mathbf{x})$ where $\mathbf{x}$ denotes the input image, and the predictive set $\{y_1,\dots, y_k\}$ obeys $0\leq k \leq K$. Therefore, $k=0$ means the recognition system rejects making a prediction, and the true label $\mathbf{y}$ is supposed to be contained in the predictive set when $\mathbf{x}$ is from known classes.

%------------------------------------------------------------------------
\section{Method}
\label{sec:method}
In this paper, we propose to tackle flexible recognition by separately estimating the confusion and ignorance for each sample. Intuitively, confusion denotes conflicting evidence between known classes, which should be additive to single-class beliefs for making reasonable multiple predictions. Ignorance denotes a lack of evidence to support rejecting samples. Most of this section falls into the proposed evidential modeling, which formulates confusion and ignorance in visual recognition under the theory of Subject Logic~\cite{josang2016subjective}. The approach to combining evidence and developing opinions follows. The approach to achieve flexible recognition is presented in the final.

\subsection{Preliminaries}
% 1. Introduce the Dempster-Shafer theory of Evidence.
%   a. Frame of discernment (k classes).
%   b. Possible hypothesis (2^k subsets, belief sum up to 1, etc.)
%   c. Formulate belief and plausibility.
Existing learning-based visual recognition models often rely on a softmax layer to give class probabilities. As a point estimation of predictive distribution, the classifier trained with the cross-entropy loss tends to deliver inflating probabilities to a single class and could not provide a reliable estimation of uncertainties~\cite{hendrycks17baseline}.

We develop confusion and ignorance based on the Dempster-Shafer Theory of Evidence (DST)~\cite{barnett2008computational}, which is a generalized scheme towards subjective probabilities~\cite{josang2016subjective}. The theory allows plausible reasoning with operations of evidence, namely the combination of evidence. Consider a $K$-class recognition task, $\Theta=\{i, 1\leq i \leq K\}$ would be the frame of discernment containing exclusive propositions, {\it{e.g.}}, class labels. Extensively, the general propositions under this frame would be the set of all subsets of $\Theta$, which is
\begin{equation}
2^{\Theta}=\{\emptyset, 1, \dots, K, \{1,2\}, \dots, \Theta\},
\end{equation}
where $2^{\Theta}$ contains a total of $2^K$ elements.

Supposing $b_A\in[0,1]$ as a measure of belief mass contained in proposition $A$, the total mass of general propositions satisfies $\sum_{A\in 2^\Theta}b_A=1$. The belief for any proposition is then defined as the summation of contained mass, which is formulated as $b_A=\sum_{B\subseteq A}b_B$.

And it is worth noting that $B\subseteq A$ suggests the logical statement that $B$ implies $A$. The total belief in $A$ is the sum of belief in all propositions that imply $A$ plus the belief in $A$ itself. For a more intuitive understanding, considering a binary visual classification task, the belief for predicting both classes is the combination of mass shared between classes and also class-exclusive masses.

We could further define the plausibility of $A$ as $pl_A$, which is the total mass of propositions that has a non-empty union with the current one. Also, the plausibility of a hypothesis could only be larger or equal to its own belief, \ie, $pl_A\geq b_A$. We demonstrate the basic probability assignments and their relations in Fig.~\ref{fig:intervals}.

\subsection{Uncertainty, Confusion, and Ignorance}
\begin{figure}[t]
\includegraphics[width=1\linewidth]{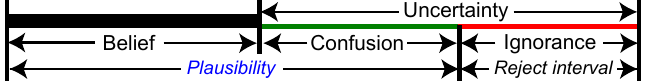}
\caption{The relation between different masses constituting the final set of opinions towards a hypothesis.}
\label{fig:intervals}
\vspace{-1mm}
\end{figure}
% 1. Introduce our definition of these two terms.
%   a. Previous works definition on uncertainties. What are their downsides?
%   b. Formulate confusion and ignorance under the Dempster-Shafer theory.
%       1) Confusion belongs to a single class.
%       2) Belief and plausibility of a single- or multi-label prediction.
%       3) Ignorance as the empty set. What does the belief of the empty set mean?
Recent evidential deep learning methods~\cite{sensoy2018evidential} develop uncertainties under multinomial opinions, which model belief for singletons as $\{b_i, i=1,\dots, K\}$ and regard the leftover as the total uncertainty $\mathcal{U}$. 
% Since the belief for a singleton only contains the probability mass of itself, the singleton belief $b_i$ for class $i$ is equal to $m_i$. 
We, therefore, have $\mathcal{U} + \sum_{i=1}^K b_i = 1$.
% These $K$ elements, together with $u$, are all non-negative and sum up to one.

However, in our modeling, we argue the uncertainty $\mathcal{U}$ for each sample $\mathbf{x}$ comes from two distinct sources, \ie, confusion $\mathcal{C}$ and ignorance $\mathcal{I}$, which is written as
\begin{equation}
\label{eq:uncertainty}
    \mathcal{U}^{\mathbf{x}}=\mathcal{C}^{\mathbf{x}}+\mathcal{I}^{\mathbf{x}}.
\end{equation}
Intuitively, a large confusion $\mathcal{C}^{\mathbf{x}}$ denotes the model is hard to distinguish $\mathbf{x}$ from known classes. For example, the model could have high confusion with a huskie image when the model has known classes of \textit{dog} and \textit{wolf}. An image from previously unknown classes, on the other hand, could have high ignorance. The superscript $\mathbf{x}$ will be omitted in the following for clarity.

To introduce the separate measurements of confusion and ignorance, our method is formalized with hyper opinions, composing masses of $2^K$ subsets for a $K$-class frame of discernment. Therefore, the overall confusion $\mathcal{C}$ is the total mass of the non-singleton subsets as $\mathcal{C}=\sum_{A,A\in 2^\Theta, 2\leq|A|\leq K} b_A$. In other words, the confusion $\mathcal{C}$ is the sum of masses shared between two or more classes. 
Altogether, following Eq.~\ref{eq:uncertainty}, we have $2^K$ mass values satisfy
\begin{equation}
    \mathcal{C} + \mathcal{I} + \sum_{i=1}^K b_i = \sum_{A,A\in 2^\Theta, 2\leq|A|\leq K} b_A + \mathcal{I}  + \sum_{i=1}^K b_i=1,
\end{equation}
where $\mathcal{I}\geq 0$ and $b\geq 0$ for all singletons and non-singleton subsets. The ignorance $\mathcal{I}$, therefore, could be regarded as the mass placed on the empty set $\emptyset$ in the frame, which indicates the level of lacking evidence. Confusion, defined on subsets with cardinalities larger than 1, reflects evidence that fails to discriminate between specific singletons. 

To further facilitate our evidence combination process, we group non-singleton confusion terms based on whether they hold evidence for a particular class $i$.

\noindent {\bf{Class $i$-related confusion mass $\mathcal{C}_i$.}} It is defined as $\mathcal{C}_i=\sum_{A, A\in 2^\Theta, |A|\geq 2,i\cap A=i} b_A$. To be more specific, $\mathcal{C}_i$ is the total mass placed on the set of all confusion terms that are supersets of singleton $i$. 
% $m_{\mathcal{C}^i}=\sum_{C\in \mathcal{C}^{\neg i}} m_C$ denotes the total confusion of classifying the sample into class $i$.

\noindent {\bf{Class $i$-unrelated confusion mass $\mathcal{C}_{\neg i}$.}} Conversely, we have $\mathcal{C}_{\neg i} = \sum_{A, A\in 2^\Theta, |A|\geq 2,i\cap A\neq i} b_A$. Accordingly, for any class $i$, we have
\begin{equation}
    \mathcal{C}=\mathcal{C}_i+\mathcal{C}_{\neg i}.
\end{equation}

And, as shown in Fig.~\ref{fig:intervals}, the plausibility of the current sample belonging to class $i$ is $pl_i=b_i+\mathcal{C}_i$, which stands for a combination of class-exclusive singleton belief mass and class-shared confusion masses.

\subsection{Evidence Combinations}
\begin{figure}[t]
\includegraphics[width=1\linewidth]{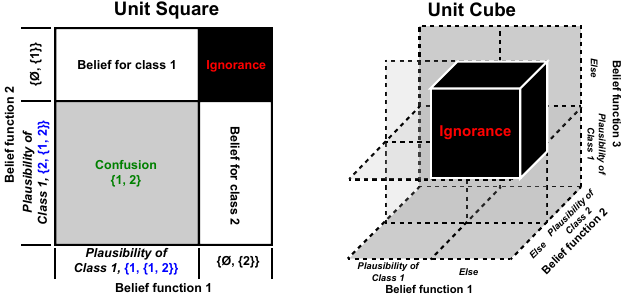}
\caption{Two graphical demonstrations of evidence combination on 2- and 3-class classification task. We only show ignorance in the 3-class example to avoid overlappings.}
\label{fig:evd_combine}
\vspace{-2mm}
\end{figure}
% 1. The rule of evidence combinations (DST).
%   a. We generalize it by giving empty-set a belief mass.
% 2. How we set up our plausibility functions.
%   a. Multiple plausibility functions under the same frame (each plausibility function sum to 1).
%   b. Class-related mass and class-unrelated mass.
%   c. Give a 3-class example (figure of a unit cube). And how we combine to form ignorance, confusion, and singleton beliefs.
%   d. Give a brief physical understanding behind the combination rule (the ratio of volume).
The objective of the proposed approach is to predict ignorance and confusion explicitly for each sample. However, unlike ignorance which is a single term, the number of confusion reaches the exponential of classes, which means our model is required to give a comprehensive quantification of $2^K$ estimates. While intimidating, we propose to handle this by decomposing the problem into $K$ plausibility functions $f_i(\cdot)$ for $i=1,\dots, K$ on the same frame. Each plausibility function $f_i(\cdot)$ is designed to give two predictions only considering class $i$, which is written as $f_i(\mathbf{x})=(pl_i, 1-pl_i)$.

Supposing we have obtained the plausibilities for all classes, any propositions, including the singleton class belief, confusion, and ignorance could then be derived by the rule of evidence combinations. In general, for $K$ plausibility functions, the belief assignment of any proposition $A$ is combined by computing as
\begin{equation} \label{eq:combine}
    b_A=\sum_{B,\cap B=A}\prod_{1\leq j\leq K} b_{B,j}(\mathbf{x})=\sum_{B,\cap B=A}\prod_{1\leq j\leq K} f_{j}^B(\mathbf{x}).
\end{equation}
Note we do not have the normalization term in our formulation because we do not exclude the empty set from our frame. We demonstrate the combination process with 2- and 3-class classification examples in Fig.~\ref{fig:evd_combine}. To clarify further, the singleton belief for class $i$ is computed as
\begin{equation}
\label{eq:belief}
    b_i=pl_i\prod_{1\leq j\leq K,j\neq i}(1-pl_j)=f_i^1(\mathbf{x})\prod_{1\leq j\leq K,j\neq i}f_j^2(\mathbf{x}).
\end{equation}
And we have the total uncertainty $\mathcal{U}=1-\sum_{i=1}^K b_i$. As the ignorance $\mathcal{I}$ could be calculated similarly as
\begin{equation}
\label{eq:ignorance}
    \mathcal{I}=\prod_{1\leq j\leq K}(1-pl_j)=\prod_{1\leq j\leq K}f_j^2(\mathbf{x}),
\end{equation}
the total confusion between all different class combinations is $\mathcal{C}=\mathcal{U}-\mathcal{I}$. The confusion term between specific classes could be further calculated with Eq.~\ref{eq:combine}.

Intuitively, the belief for each combination could be regarded as the occupation in a unit $K$-dim volume. The modeling of confusion becomes feasible by spanning a $K$-dim hypothesis space with $K$ plausibility functions. Moreover, the computation complexity of each combination is $\mathcal{O}(n)$, and for specific conditions, we could only calculate necessary confusion terms.

\subsection{Developing Opinions}
% 1. We model each plausibility function as a binary prediction.
%   a. Two different approaches, k-binary or sigmoid-1 linear layer.
% 2. Major loss, the same formulation with EDL.
%   a. Dirichlet distribution.
%       1) The formulation of Dirichlet distribution on categorical distribution.
%       2) The prior parameter is for our singleton belief. 
%   b. Using NLL loss to train our model.
% 3. Our regularization term.
%   a. The problem with using major loss only. Could obtain undesired results.
%   b. The formulation of our regularization term.
% 4. KL loss.
%   a. The motivation of adding KL loss. Encourage singleton belief after combination.

% estimating mass on empty set is not possible
In this section, we describe how to develop opinions from training data. Each plausibility function $f_i(\cdot)$ can be constructed as a normalized dual-output linear layer or a single multi-output layer after being activated by a sigmoid function $\sigma(\cdot)$. In this work, the second is implemented to reduce the network parameters. In particular, the output is regarded as the value of class plausibility. The plausibility function is then formulated as
\begin{equation}
    (pl_i, 1-pl_i)=f_i(\mathbf{x})=\sigma(w^{\top}_i\Phi(\mathbf{x}))
\end{equation}
where $\Phi:\mathcal{X}\rightarrow\mathbb{R}^{D}$ is a feature embedding function.

Typically, only one deterministic label is given for each image in a standard visual recognition dataset. 
Following EDL~\cite{sensoy2018evidential}, the learning of singleton belief is implemented as evidence acquisition on a Dirichlet prior. The loss of EDL is
% We assume the belief for singletons follow a prior Dirichlet distribution as in EDL~\cite{sensoy2018evidential}, which indicates the learning has transformed from maximizing class multinomial likelihoods to estimating the parameter $\alpha$ of Dirichlet distribution. 
%$\alpha$ could be computed as $e+1$, where $e$ is the class-exclusive evidence. 
% The loss to be minimized for a sample $x_t$ after integrating out the class probabilities is
\begin{equation}
    % \mathcal{L}^t_{EDL}=\sum^K_{i=1}y_{ti}(\log(S_t)-\log(\alpha_{ti})),
    \mathcal{L}_{\text{EDL}}=\sum^K_{i=1}y_{i}[\log(\sum^K_{j=1}\alpha_{j})-\log(\alpha_{i})],
\end{equation}
where $\mathbf{y}=[y_1, y_2, \ldots, y_{i}, \ldots, y_{K}]^T$ is one-hot class label for a sample $\mathbf{x}$, and $\boldsymbol\alpha=[\alpha_1, \alpha_2, \ldots \alpha_{K}]^T$ are parameters of a Dirichlet distribution $Dir(\cdot|\boldsymbol\alpha)$. 

Different from EDL, class evidence is replaced with belief. Hence, $\boldsymbol\alpha$ is directly calculated from singleton beliefs and overall uncertainty. It is derived as
% During training, we calculate $\alpha_{i}$ by first calculating the total uncertainty as $u=1-\sum_{j=1}^Km_{j}$, where the singleton beliefs $m_{j}$ are obtained by (\ref{eq:combine}). The $\alpha_{i}$ is computed as
\begin{equation}
    % \alpha_i=b_i S+1,
    \alpha_i=\frac{K b_i}{\mathcal{U}}+1=\frac{K b_i}{1-\sum_{j=1}^Kb_{j}}+1,
\end{equation}
where $b_{i}$ could be obtained from Eq.~\ref{eq:belief}.
During inference, all opinions could be directly predicted by performing combinations on the output of plausibility functions.

To encourage the plausibility function to match our expected behavior, \ie, predicting the plausibility instead of the belief of singleton, we add a regularization term as
\begin{equation}
\label{eq:reg}
    \mathcal{L}_{\text{reg}}=\sum^K_{i=1}y_{i}[pl_i-(1-\hat{\mathcal{I}})]^2,
\end{equation}
where $\hat{\mathcal{I}}$ is the current estimation of ignorance. 

Following EDL, a Kullback-Leibler loss is used to minimize evidence on unrelated classes as
% Moreover, we minimize the Kullback-Leibler (KL) divergence between the calculated evidence and a uniform Dirichlet distribution to motivate learning class-exclusive evidence for singletons. The KL term is formulated as
\begin{equation}
\label{eq:kl}
    \mathcal{L}_{\text{KL}}=\textit{KL}(Dir(\cdot|\tilde{\boldsymbol\alpha})||Dir(\cdot|\langle 1, \dots, 1\rangle)),
\end{equation}
where $\tilde{\boldsymbol\alpha}=\mathbf{y}+(1-\mathbf{y})\odot\boldsymbol\alpha$, $\odot$ for element-wise multiplication. Combining all terms together yields the final loss as
\begin{equation}
\label{eq:loss_all}
    \mathcal{L} = \mathcal{L}_{\text{EDL}} + \lambda_{\text{reg}} \mathcal{L}_{\text{reg}} + \lambda_{\text{KL}} \mathcal{L}_{\text{KL}}.
\end{equation}
Each loss term is accompanied by a balance weight, and we gradually increase the effect of $\mathcal{L}_{kl}$ through an additional annealing coefficient.

After developing opinions with the proposed method, a straightforward solution would be setting the belief threshold for outputs to achieve flexible recognition. The sample will be rejected if the ignorance is too large that no combination would exceed the threshold. And the model gives incrementally combined predictions if no singleton belief meets the bar.

%------------------------------------------------------------------------
\section{Experiments}
\label{sec:experiments}

The proposed method could model two sources of uncertainty for each sample to handle the task of flexible recognition. To enable better comparison with existing methods, our experiments are primarily decomposed into three components. (1) Demonstrating the separation of two sources of uncertainty, \ie, confusion and ignorance. (2) Indicating the correct class on misclassified samples with estimated confusion. 
% The designed evaluation metric could avoid the impact of different sizes of considered sets. 
(3) Applying ignorance to compare with other methods on the task open-set detection. More experiments, including on adversarial-attacked samples and ablation studies, follow to give a more comprehensive evaluation.

\noindent{\bf{Implementation details.}} We adopt the ResNet-18 as the backbone for our experiments except on synthetic data and open-set detection. The dimension of extracted feature is set to $512$. For the proposed method, we apply the sigmoid activation on the last linear layer to work as our multiple plausibility functions. 
We empirically find both EDL~\cite{sensoy2018evidential} and our method are more sensitive to the learning rate. Specifically, we set the learning rate for both methods to $0.004$ with a momentum of $0.9$ for the batch size of $128$. 
$\lambda_{\text{KL}}$ in Eq.~\ref{eq:loss_all} anneals to $0$ with epochs with the maximum coefficient of $0.05$, and $\lambda_{\text{reg}}$ is set to $1$. 
% For experiments on CIFAR-100~\cite{krizhevsky2009learning}, we use the binary cross-entropy loss to facilitate the training of the proposed method in the first 5 epochs. The reason is the cold start problem as 

\subsection{Synthetic Experiments}
\begin{figure}[t]
\centering
    \begin{subfigure}[b]{0.326\linewidth}
    \centering
    \includegraphics[width=\linewidth]{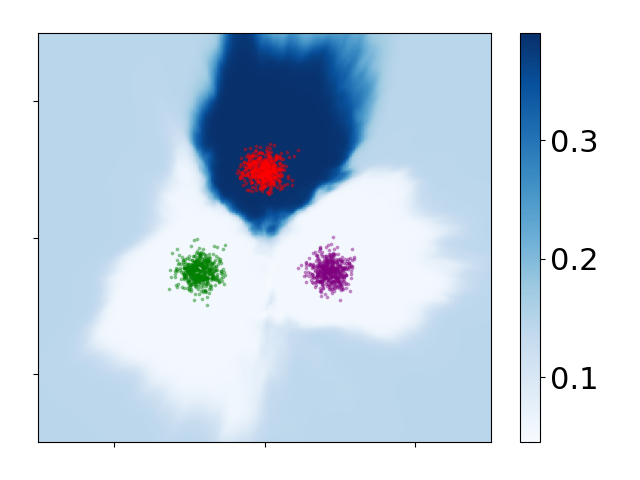}
    \caption{Class 1}
    \end{subfigure}
    \begin{subfigure}[b]{0.326\linewidth}
    \centering
    \includegraphics[width=\linewidth]{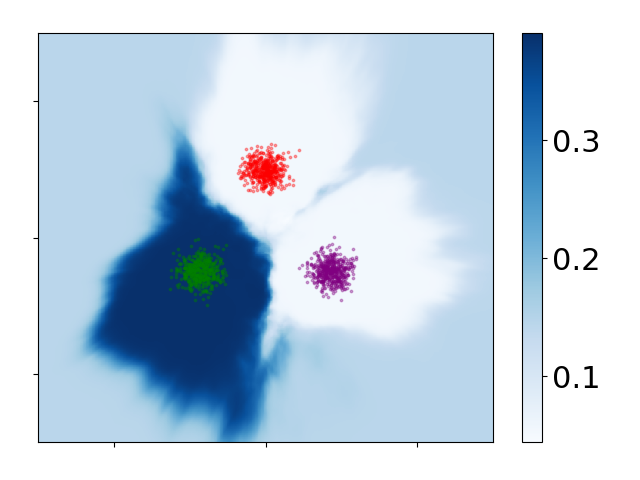}
    \caption{Class 2}
    \end{subfigure}
    \begin{subfigure}[b]{0.326\linewidth}
    \centering
    \includegraphics[width=\linewidth]{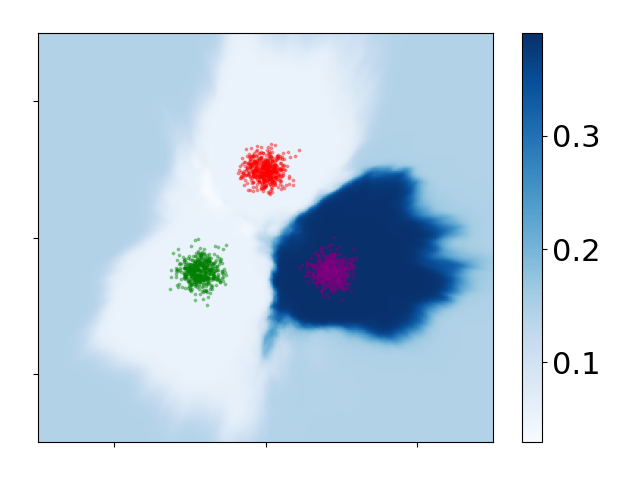}
    \caption{Class 3}
    \end{subfigure}
    \begin{subfigure}[b]{0.326\linewidth}
    \centering
    \includegraphics[width=\linewidth]{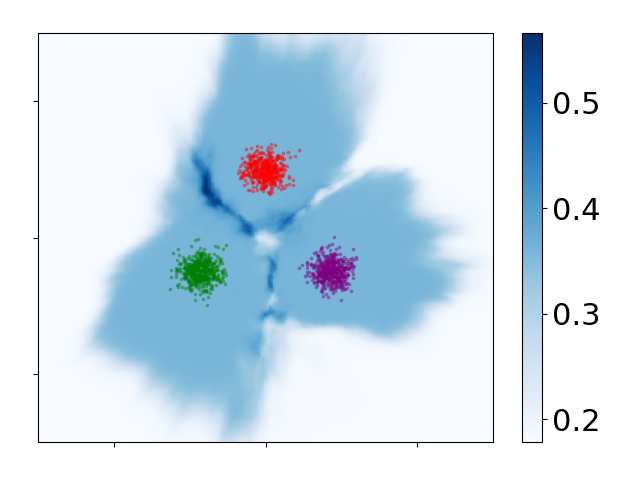}
    \caption{Confusion}
    \end{subfigure}
    \begin{subfigure}[b]{0.326\linewidth}
    \centering
    \includegraphics[width=\linewidth]{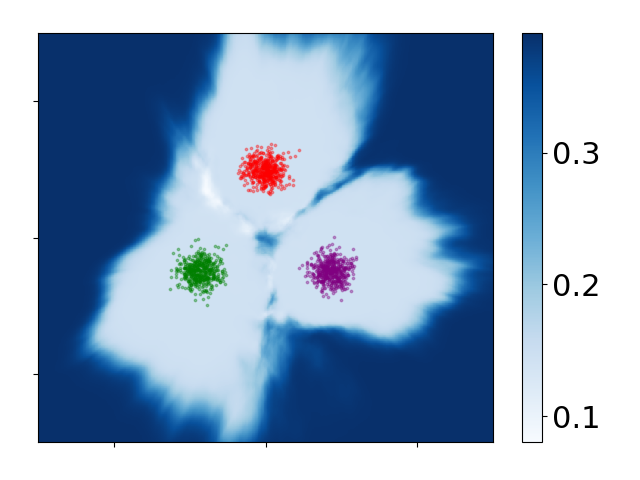}
    \caption{Ignorance}
    \end{subfigure}
    \begin{subfigure}[b]{0.326\linewidth}
    \centering
    \includegraphics[width=\linewidth]{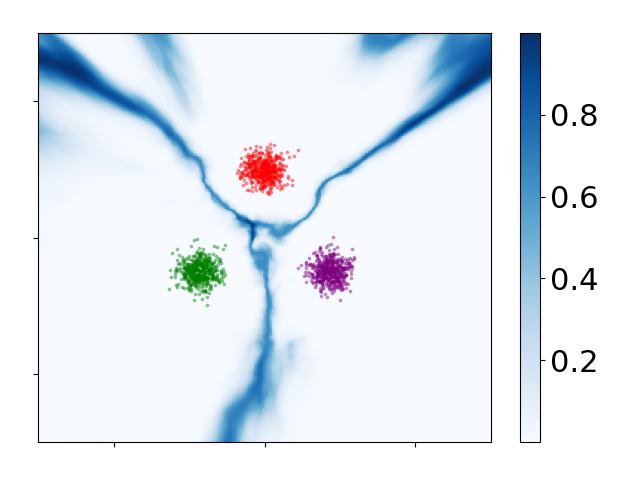}
    % \caption{Entropy (Standard)}
    \caption{Standard   Entropy}
    \end{subfigure}
\caption{A 3-class classification problem. The Gaussian-distributed training data are depicted with dots, while the background color indicates the estimated value of the corresponding location. Our results are plotted from (a) to (e). The entropy over predictions of a standard net trained with cross-entropy loss is shown in (f) for comparison.}
\label{fig:synthetic}
\vspace{-2mm}
\end{figure}
% 1. To show the behavior of different methods in a low-dim space.
%   Advantageous of our method in this section:
%   a. could separate uncertainty into confusion and ignorance.
%   b. the belief area for each class is more compact.
%   c. purer ignorance could benefit open-set detection.
%   d. the confusion gives a better understanding of the current sample. 
%   Baselines:
%   a. Max-entropy. b. Bayesian dropout. c. Deep ensemble method.
%   Data:
%   a. 3 Gaussian. b. More complex data (like circles, ovals, or other inseparable data).

% As confusion happens among all class combinations, to avoid combinatorial complexity, plausibility functions are introduced to predict per-class evidence for each image. The confusion could be directly inferred through evidence combinations. Fig.~\ref{fig:firstsight_b} compares the behavior of our approach with the entropy of a standard network in the 2-dimensional space. As we can observe, the proposed method acts as a density estimator. Confusion happens between class boundaries, while ignorance is high for out-of-distribution data points. In addition, the proposed method does not require intricate adaptions to network architectures to achieve the desired properties.

In this section, we examine the behavior of our approach in Fig.~\ref{fig:synthetic} in the 2-dimensional space. A dataset with three isotropic Gaussian distributed classes is created for training. The distances between each Gaussian are equal to $9$, and we set $\sigma=4$ for all Gaussian. Each class has $500$ training samples which are denoted by colored dots. We discretize the background into 2D locations for testing. The background color denotes corresponding estimated values.

Both our method and the standard entropy are trained with the same multi-layer perceptron. As we can observe in Fig.~\ref{fig:synthetic} (a) to (e), the proposed method acts as a density estimator. Confusion happens between class boundaries, while ignorance is high for out-of-distribution data points. The proposed method does not require intricate adaptions to network architectures to achieve the desired properties. In Fig.~\ref{fig:synthetic} (f), the entropy of a standard deterministic neural net with the cross-entropy loss is plotted, which shows that using entropy to distinguish out-of-distributions could face multiple downsides due to its sharp boundaries.

\subsection{Confusion on Misclassified Samples}
\label{sec:confusion1}

\begin{table}[t]
\scriptsize
\centering
% \begin{tabular}{c|ccc|ccc}
% \hline
% \multirow{3}{*}{\textbf{Dataset}} & \multicolumn{3}{c|}{\textbf{CIFAR-10}} & \multicolumn{3}{c}{\textbf{CIFAR-100}} \\ \cline{2-7} 
%  & \multicolumn{1}{c|}{\textbf{All}} & \multicolumn{2}{c|}{\textbf{Misclassified}} & \multicolumn{1}{c|}{\textbf{All}} & \multicolumn{2}{c}{\textbf{Misclassified}} \\ \cline{2-7} 
%  & \multicolumn{1}{c|}{Acc.} & \multicolumn{1}{c|}{AUPR} & AUROC & \multicolumn{1}{c|}{Acc.} & \multicolumn{1}{c|}{AUPR} & AUROC \\ \hline\hline
% CE-L2 & \multicolumn{1}{c|}{95.17} & \multicolumn{1}{c|}{27.03} & 63.40 & \multicolumn{1}{c|}{76.01} & \multicolumn{1}{c|}{4.26} & 57.26 \\ \hline
% EDL~\cite{sensoy2018evidential} & \multicolumn{1}{c|}{94.82} & \multicolumn{1}{c|}{22.74} & 60.41 & \multicolumn{1}{c|}{74.50} & \multicolumn{1}{c|}{3.27} & 54.90 \\ \hline
% Dropout~\cite{gal2015bayesian} & \multicolumn{1}{c|}{94.72} & \multicolumn{1}{c|}{31.87} & 64.53 & \multicolumn{1}{c|}{74.49} & \multicolumn{1}{c|}{6.74} & 58.91 \\ \hline
% One-vs-Rest & \multicolumn{1}{c|}{95.07} & \multicolumn{1}{c|}{26.70} & 63.21 & \multicolumn{1}{c|}{72.22} & \multicolumn{1}{c|}{8.15} & 63.35 \\ \hline
% ASL~\cite{ben2020asymmetric} & \multicolumn{1}{c|}{95.16} & \multicolumn{1}{c|}{34.27} & 70.91 & \multicolumn{1}{c|}{75.80} & \multicolumn{1}{c|}{14.62} & 79.96 \\ \hline
% Ours & \multicolumn{1}{c|}{94.96} & \multicolumn{1}{c|}{\textbf{65.01}} & \textbf{89.51} & \multicolumn{1}{c|}{74.92} & \multicolumn{1}{c|}{\textbf{31.42}} & \textbf{90.01} \\ \hline
% \end{tabular}
\begin{tabular}{c|cc|cc|cc}
\hline
\multirow{2}{*}{\textbf{Dataset}} & \multicolumn{2}{c|}{\textbf{CIFAR-10}} & \multicolumn{2}{c|}{\textbf{CIFAR-100}} & \multicolumn{2}{c}{\textbf{Imagenet}} \\ \cline{2-7} 
 % & \multicolumn{1}{c|}{\textbf{All}} & \textbf{Mis.} & \multicolumn{1}{c|}{\textbf{All}} & \textbf{Mis.} & \multicolumn{1}{c|}{\textbf{All}} & \textbf{Mis.} \\ \cline{2-7} 
 & \multicolumn{1}{c|}{Acc.} & AUROC & \multicolumn{1}{c|}{Acc.} & AUROC & \multicolumn{1}{c|}{Acc.} & AUROC \\ \hline\hline
Softmax & \multicolumn{1}{c|}{95.2} & 63.4 & \multicolumn{1}{c|}{76.0} & 57.3 & \multicolumn{1}{c|}{54.3} & 58.2 \\ \hline
EDL~\cite{sensoy2018evidential} & \multicolumn{1}{c|}{94.8} & 60.4 & \multicolumn{1}{c|}{74.5} & 54.9 & \multicolumn{1}{c|}{54.2} & 58.3 \\ \hline
Dropout~\cite{gal2015bayesian} & \multicolumn{1}{c|}{94.7} & 64.5 & \multicolumn{1}{c|}{74.5} & 58.9 & \multicolumn{1}{c|}{47.3} & 61.8 \\ \hline
OvR & \multicolumn{1}{c|}{95.1} & 63.2 & \multicolumn{1}{c|}{72.2} & 63.4 & \multicolumn{1}{c|}{46.3} & 60.2 \\ \hline
ASL~\cite{ben2020asymmetric} & \multicolumn{1}{c|}{95.2} & 70.9 & \multicolumn{1}{c|}{75.8} & 79.9 & \multicolumn{1}{c|}{54.1} & 62.8 \\ \hline
Ours & \multicolumn{1}{c|}{95.0} & \textbf{89.5} & \multicolumn{1}{c|}{74.9} & \textbf{90.0} & \multicolumn{1}{c|}{54.6} & \textbf{97.6} \\ \hline
\end{tabular}
\caption{Comparison on whether the confusion indicates the correct class on misclassified samples. Results of classification accuracy and AUROC on misclassified samples are shown on three datasets with different class scales.}
\label{tab:confusion}
\vspace{-2mm}
\end{table}

% the Area Under the Precision-Recall curve (AUPR) and
In this part, the confusion is tested on misclassified samples, checking whether it is correlated with the ground-truth label. That is, the confusion should be high between the misclassified class and the ground-truth class. Besides in-sample comparison on the level of confusion, we argue the between-sample comparability of confusion is also critical for flexible visual recognition. For example, a flexible recognition system would tend to give a second prediction if its confusion is higher than in other samples. To address this concern, we employ the Area Under the Receiver Operating Characteristic curve (AUROC) as our evaluation metric, which sorts the predicted value along samples in each class. More specifically, the ROC curve is a graph showing the true positive rate against the false positive rate. A random classifier would correspond to a $50\%$ AUROC. Moreover, the metric sidesteps the issue of threshold selection. 

To be more clear, the confusion terms between all classes and the predicted class are regarded as the input to the metrics, while the target is one-hot class labels. As the ground-truth class could be imbalanced among misclassified samples, the AUROC is weighted by the categorical base rate.

The results on CIFAR-10, CIFAR-100~\cite{krizhevsky2009learning}, and ImageNet~\cite{deng2009imagenet} with 10, 100, and 1000 classes, respectively, are shown in Tab.~\ref{tab:confusion} to demonstrate the comparison with different class scales. The Imagenet used here is an official downsampled version with the resolution of $64\times 64$. Besides standard Softmax, the Dropout~\cite{gal2015bayesian} refers to the method using Monte Carlo Dropout (with the dropout rate of 0.2 and 10 dropout iterations) to approximate the
distribution of model parameters. Since these methods do not contain explicit outputs of confusion, they are compromised to an intuitive approach, which measures the difference between predicted class probabilities. Two multi-label classification methods, namely, the One-vs-Rest (OvR) and the ASL~\cite{ben2020asymmetric}, are included as baselines whose outputs are not normalized between classes. For EDL~\cite{sensoy2018evidential}, confusion is defined as the difference between estimated singleton beliefs. In stark contrast to these methods, the proposed method uses explicitly estimated confusion between any classes with the predicted class following the definition in Eq.~\ref{eq:combine}.

As shown in Tab.~\ref{tab:confusion}, the performance of the proposed method is significantly higher than all other approaches on three datasets. There are two reasons to support this result. First, the probability produced by softmax should not be viewed as the direct measure of confidence for each class~\cite{hendrycks2016baseline}, which means there might be no suitable way to derive confusion from softmax probability after training. Second, for the evidential method~\cite{sensoy2018evidential}, the difference between predicted beliefs does not capture their shared features. In other words, the confusion in EDL is mixed with ignorance in their one uncertainty estimate. Orthogonal to these methods, the proposed approach could model the confusion that occurred between classes and could help flexible recognition systems determine when and which class to be predicted next. 
% We also provide the results with the metric of the Area Under the Precision-Recall curve (AUPR) in the supplementary.

\subsection{Flexible Closed-set Recognition}
\begin{figure}[t]
\centering
    \begin{subfigure}[b]{1\linewidth}
    \centering
    \includegraphics[width=\linewidth]{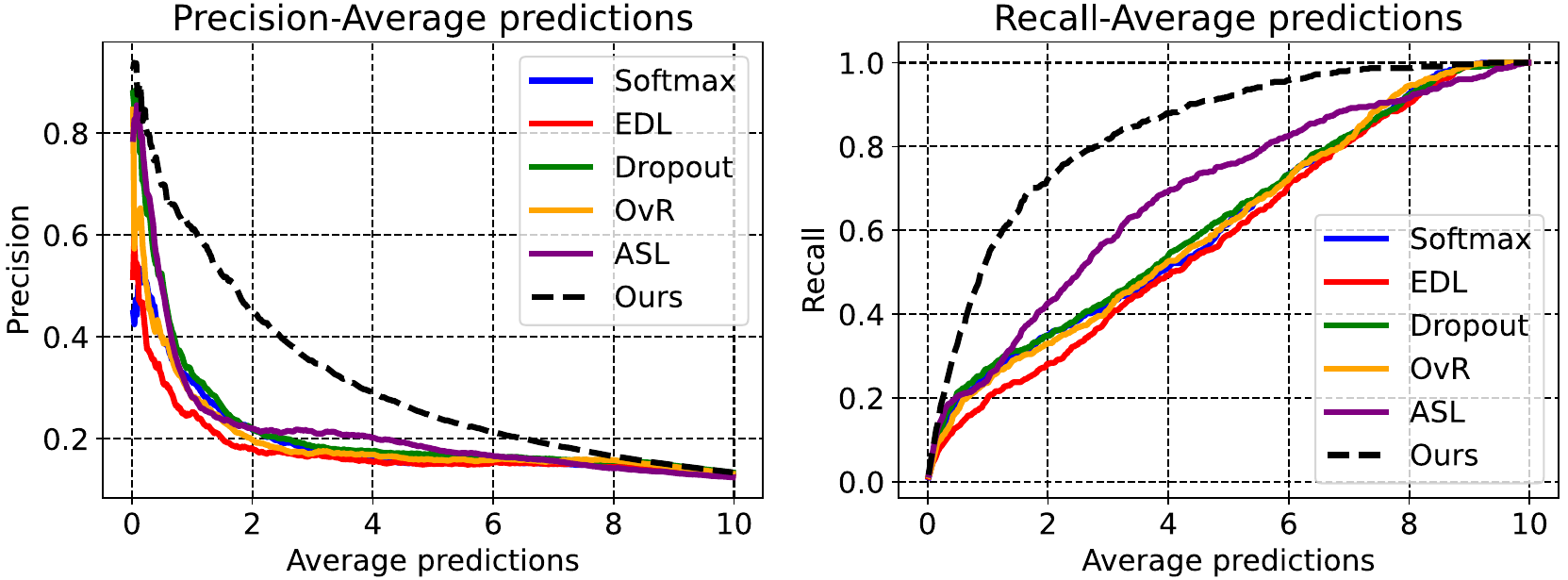}
    \caption{Results on CIFAR-10}
    \label{fig:curve:cifar10}
    \end{subfigure}
    \begin{subfigure}[b]{1\linewidth}
    \centering
    \includegraphics[width=\linewidth]{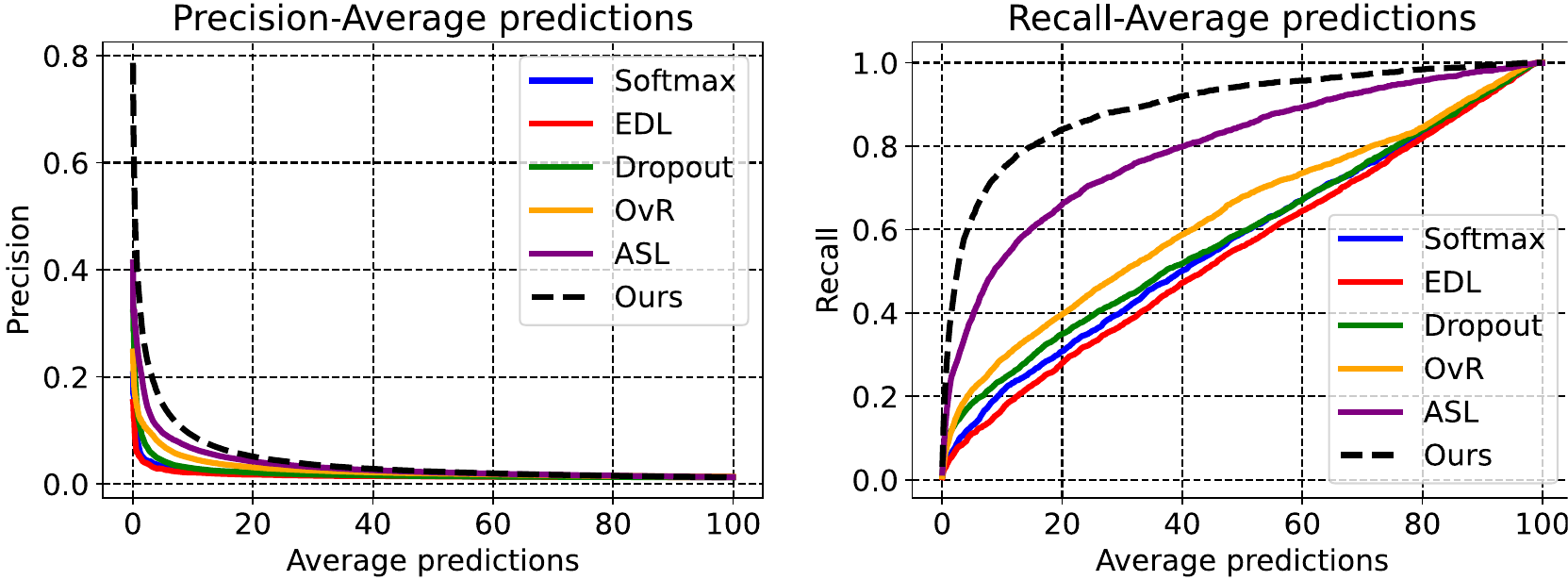}
    \caption{Results on CIFAR-100}
    \label{fig:curve:cifar100}
    \end{subfigure}
    
    \begin{subfigure}[b]{1\linewidth}
    \centering
    \includegraphics[width=\linewidth]{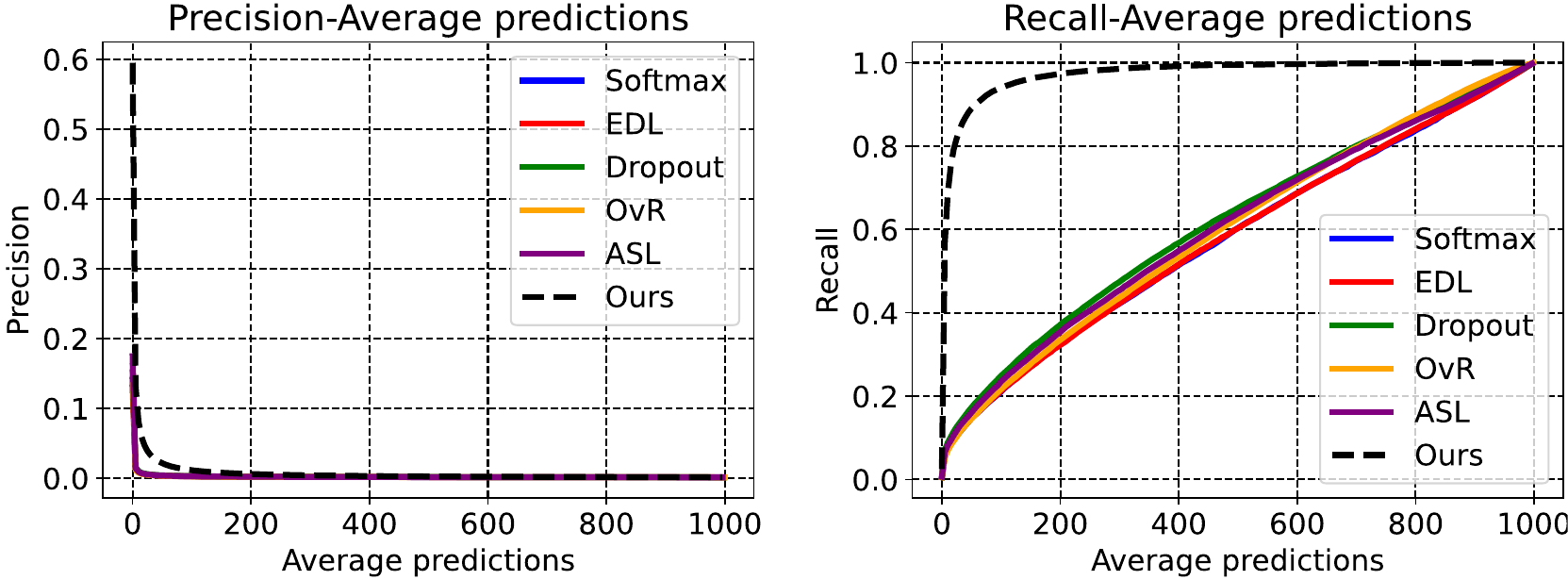}
    \caption{Results on Imagenet}
    \label{fig:curve:imagenet}
    \end{subfigure}
\caption{Precision and recall of delivering multiple predictions on misclassified samples with respect to the average number of predictions. The intention to predict extra classes is dependent on the confusion term between classes.}
\label{fig:curve}
\vspace{-2mm}
\end{figure}

We turn to the task of flexible visual recognition on the closed CIFAR-10, CIFAR-100, and Imagenet, where methods are supposed to make multiple predictions when they are unsure. And we regard the output as correct as long as the correct answer is included in the prediction set. However, simply increasing the number of predictions for each sample should be penalized. Thus, for a more meaningful comparison, we show precision and recall as the function of the average number of predictions. We only evaluate methods on their misclassified samples, where providing more predictions is urgently demanded. And the indicator for making another prediction is the confusion between the considered and the predicted class, the same as in Section~\ref{sec:confusion1}. The confusion for the first-predicted class itself is intuitively described as $0$.

The curves of three datasets are drawn in Fig.~\ref{fig:curve}. We incrementally select more predictions for each class, \ie, the samples are deserved to make an extra prediction if its confusion on this considered class is more significant than in other samples. By doing this, we could control the average predictions while being threshold-independent in evaluating flexible visual recognition. A desired precision-average prediction curve would have a near $1$ value for the start of making predictions. For the recall-average prediction curve, the closer it adheres to the left-top corner, the earlier it delivers the correct class. When comparing the precision curve, our method could achieve a significantly higher value than other methods when making limited predictions. For example, on the CIFAR-10 dataset, the proposed method reaches the precision of $0.62$ for making an average of only one prediction for each sample, while the second highest is $0.33$. Note some samples could have no prediction at the beginning as all of their confusion terms are lower than other samples. Besides, we also notice that ASL~\cite{ben2020asymmetric} is better than other baselines in the recall curve. This indicates that for flexible recognition, a multi-label classifier could be a better choice as it prevents the overconfident problem in the softmax probabilities to some extent.

\subsection{Ignorance for Open-set Detection}
% \begin{table}[t]
% \scriptsize
% \caption{Comparison with other methods on open-set detection. Note the Acc. here stands for open-set detection accuracy, and the F1 score is the Macro-F1, which evaluates $K+1$ classes.}
% \label{tab:open}
% \begin{tabular}{c|ccc|ccc}
% \hline
% \multirow{2}{*}{\textbf{Dataset}} & \multicolumn{3}{c|}{\textbf{CIFAR-10 + LSUN (crop)}} & \multicolumn{3}{c}{\textbf{CIFAR-10 + ImageNet (crop)}} \\ \cline{2-7} 
%  & \multicolumn{1}{c|}{Acc.} & \multicolumn{1}{c|}{F1} & AUROC & \multicolumn{1}{c|}{Acc.} & \multicolumn{1}{c|}{F1} & AUROC \\ \hline\hline
% CE-L2 & \multicolumn{1}{c|}{70.37} & \multicolumn{1}{c|}{0.842} & 95.58 & \multicolumn{1}{c|}{65.81} & \multicolumn{1}{c|}{0.823} & 95.24 \\ \hline
% Entropy & \multicolumn{1}{c|}{73.48} & \multicolumn{1}{c|}{0.851} & 96.14 & \multicolumn{1}{c|}{68.85} & \multicolumn{1}{c|}{0.833} & \textbf{95.79} \\ \hline
% EDL~\cite{sensoy2018evidential} & \multicolumn{1}{c|}{71.96} & \multicolumn{1}{c|}{0.855} & 95.45 & \multicolumn{1}{c|}{70.29} & \multicolumn{1}{c|}{0.834} & 95.87 \\ \hline
% Dropout~\cite{gal2015bayesian} & \multicolumn{1}{c|}{52.55} & \multicolumn{1}{c|}{0.765} & 92.46 & \multicolumn{1}{c|}{46.90} & \multicolumn{1}{c|}{0.745} & 90.90 \\ \hline
% Ours & \multicolumn{1}{c|}{\textbf{79.86}} & \multicolumn{1}{c|}{\textbf{0.869}} & \textbf{97.03} & \multicolumn{1}{c|}{\textbf{71.65}} & \multicolumn{1}{c|}{\textbf{0.839}} & 95.76 \\ \hline
% \end{tabular}
% % \vspace{-2mm}
% \end{table}

\begin{table}[t]
\scriptsize
\centering
\begin{tabular}{c|c|c}
\hline
\textbf{Closed Dataset CIFAR-10} & \textbf{+ LSUN (crop)} & + \textbf{ImageNet (crop)} \\ \hline\hline
Softmax & 64.2 & 63.9 \\ \hline
OpenMax~\cite{bendale2016towards} & 65.7 & 66.0 \\ \hline
OSRCI~\cite{neal2018open} & 65.0 & 63.6 \\ \hline
LadderNet + OpenMax~\cite{yoshihashi2019classification} & 65.2 & 65.3 \\ \hline
DHRNet + OpenMax~\cite{yoshihashi2019classification} & 65.6 & 65.5 \\ \hline
CROSR~\cite{yoshihashi2019classification} & 72.0 & 72.1 \\ \hline
GFROSR~\cite{perera2020generative} & 75.1 & 75.7 \\ \hline
% EDL~\cite{sensoy2018evidential} & 76.3 & 72.8 \\ \hline
Ours & \textbf{80.5} & \textbf{76.8} \\ \hline
\end{tabular}
\caption{Comparison on open-set detection by adding different unknown samples to the test set. The performance is evaluated by Macro-F1, which considers $K+1$ classes.}
\label{tab:open}
\vspace{-2mm}
\end{table}
% Metric:
% Costs of mis-decisions
% Proper scoring rules are scoring rules that are minimized in expectation if the predictive density is the true density. 
% 1. Dataset: 
%   a. Closed: CIFAR-10, CIFAR-100, CUB
%   b. Open: ImageNet, LSUN, iSUN (crop & resize)
% 2. Metric, explain the reason why we choose these metrics:
%   a. Closed-set classification acc. for reference.
%   b. 3-class classification (closed and correct, closed but misclassified, open).
%       (1) Macro-AUPR (2) OVO AUROC
%       We could separate.
%   c. n-classes confusion value test.
%       (1) weighted AUPR (2) weighted AUROC
%       Our binary class-wise confusion correctly reflects the model's limitations when misclassification happens.
%   d. Open-set detection
%       (1) Open-set detection ACC. (2) Macro F1
%       Compared to EDL, our ignorance is more suitable.
% 3. Baselines:
%   a. Base random model. We show what a random model could achieve.
%   b. Other uncertainty estimation methods.

% \subsection{Open-set and Adversarial Disturbance}
As a crucial part of flexible recognition, we additionally demonstrate the effectiveness of our method on the task of open-set detection in Tab.~\ref{tab:open} following standard protocols~\cite{yoshihashi2019classification,hendrycks2016baseline,liang2017enhancing}. The same network architecture, \ie, a 13-layer VGG model, is used to implement the proposed method as in~\cite{yoshihashi2019classification}. 
% For the metric of open-set detection accuracy and AUROC, closed-set samples are considered as negative samples, while open-set samples constitute positive samples. 
% Methods are conducted with the same training protocol and architectures as described in our previous CIFAR-10 experiments. And no open-set data is included during the training process.

During the test time of open-set detection, the test images from CIFAR-10 datasets are viewed as closed-set examples. For open-set samples, we consider two natural image datasets, \ie, LSUN (crop) and ImageNet (crop), introduced by Liang et al~\cite{liang2017enhancing}. The Macro-F1 is therefore evaluated on $K+1$ classes by regarding all open samples as an additional class. And the threshold for being detected as open-set is chosen when $95\%$ of closed-set images can be correctly classified. Among the compared methods, OSRCI~\cite{neal2018open} augments the training set with hard and counterfactual generated images to improve unknown sample detection during testing. CROSR~\cite{yoshihashi2019classification} and GFROSR~\cite{perera2020generative} incorporate reconstruction loss into the procedure of closed-set training to better model the data distribution. For the proposed method, we use ignorance $\mathcal{I}$ as the indicator of unknown samples and achieve better performance without additional adversarial training data. The F1 score of the proposed method is higher than other methods on both datasets, which supports the claim that ignorance in the proposed method could effectively handle the lack of evidence in open-set samples. It is worth noting that, besides explicitly delivering ignorance to detect unknown samples, the advantage of the proposed method also lies in the estimate of confusion, which is evaluated in closed-set experiments.

% is stemmed from previous uncertainty estimation methods where we believe confusion and ignorance are better individually quantified. Accordingly, the advantage of our method is established on the successful estimation of confusion and ignorance instead of outperforming state-of-the-art OOD methods, especially for those that do not explicitly deliver uncertainties. 
% As the proposed method lies in the domain of uncertainty quantification, we compare the proposed method to other similar methods. As our method could separate confusion and ignorance, we only use the estimate of ignorance for detecting open-set samples, which is supposed to describe the lack of evidence, \ie, the degree of out-of-distribution. Compared to EDL~\cite{sensoy2018evidential}, the open-set detection accuracy of the proposed method achieves the absolute increase of $7.9\%$ and $1.36\%$ on both datasets. More importantly, we demonstrate the effectiveness of the proposed method in rejecting samples when performing flexible recognition.

\subsection{Performance on Adversarial Samples}
\begin{figure}[t]
\centering
    \centering
    \includegraphics[width=\linewidth]{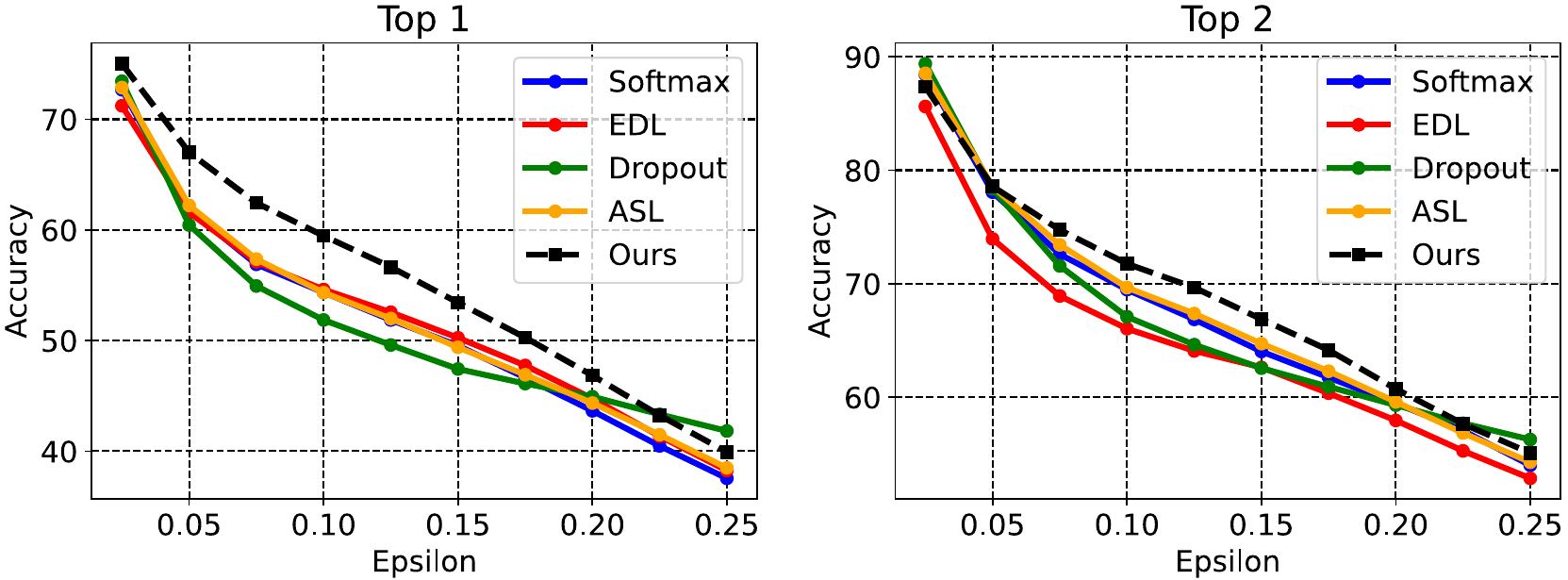}
\caption{Top 1 and 2 accuracies with adversarial perturbations $\epsilon$ on CIFAR-10~\cite{krizhevsky2009learning} dataset.}
\label{fig:adversarial}
\vspace{-2mm}
\end{figure}
We compare the robustness of different classification methods~\cite{gal2015bayesian,sensoy2018evidential} against adversarial attacks. Besides models introduced in the previous experiment, we include the recent multi-label classification method ASL~\cite{ben2020asymmetric} as another baseline. The reason is that multi-label classification methods usually adopt the same sigmoid activation and multiple binary linear layers as our method. For our method, we use the plausibility $pl_i$ for class $i$ to predict instead of their singleton beliefs, \ie, the confusion for each class that occurred during adding perturbations is involved in the ranking. The reason is that we want to test whether the confusion correctly characterizes the conflicting evidence shared between the correct and other classes. 
% Please refer to Tab.~\ref{tab:confusion} for the closed-set testing accuracies without perturbations. 

Adversarial samples are generated on CIFAR-10~\cite{krizhevsky2009learning} using the Fast Gradient Sign method~\cite{goodfellow2016cleverhans} with various perturbation parameters $\epsilon$. The adversarial attack method uses the gradient during inference to generate samples that are more challenging to make correct predictions as $\epsilon$ increases. Both the top 1 and top 2 results are shown in Fig.~\ref{fig:adversarial}. The figure indicates that the proposed method could almost achieve both the highest top 1 and top 2 results. That is, the confusion that happened during perturbating is captured by our method, which could still contribute to the corresponding correct class. The proposed method is only slightly worse than the Dropout method when $\epsilon=0.25$, which is forgivable as its training includes stochastic zeroing on the model parameter to increase its robustness.

\begin{figure*}[t]
\centering
    \begin{subfigure}[b]{0.32\linewidth}
    \centering
    \includegraphics[width=\linewidth]{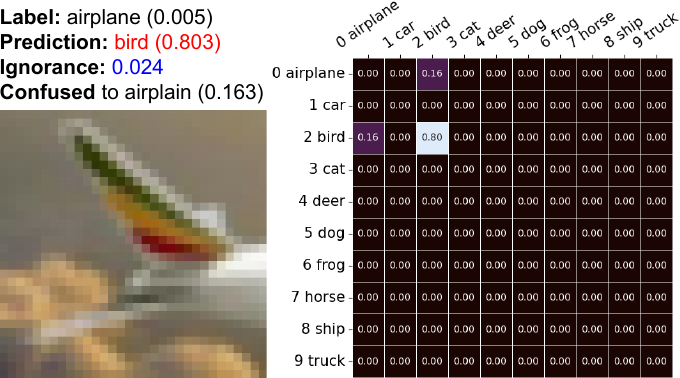}
    \caption{}
    \label{fig:qua_1}
    \end{subfigure}
    \begin{subfigure}[b]{0.32\linewidth}
    \centering
    \includegraphics[width=\linewidth]{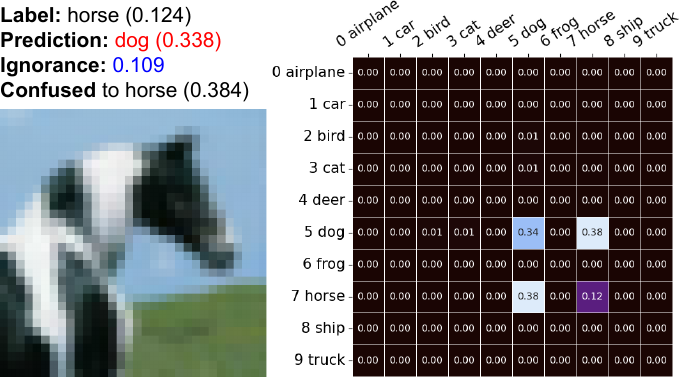}
    \caption{}
    \label{fig:qua_2}
    \end{subfigure}
    \begin{subfigure}[b]{0.32\linewidth}
    \centering
    \includegraphics[width=\linewidth]{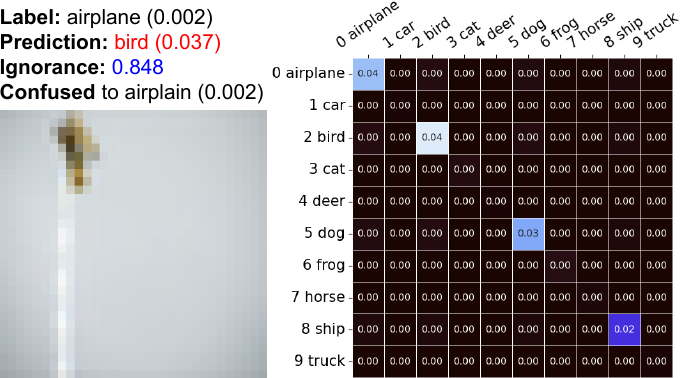}
    \caption{}
    \label{fig:qua_6}
    \end{subfigure}
\caption{Matrices of confusion of misclassified samples on the CIFAR-10 dataset. The diagonal of each matrix is set to the singleton belief of each class. Notice that the background color for each heatmap is normalized individually. The total ignorance is demonstrated in the caption.}
\label{fig:qualitive}
\vspace{-2mm}
\end{figure*}
\subsection{Qualitative Results}
% 1. Per-sample matrix of confusion and ignorance.
Furthermore, to provide a more intuitive impression of our results, we demonstrate the matrics of binary confusion estimates in Fig.~\ref{fig:qualitive}. We take the misclassified images of the proposed method for the CIFAR-10 dataset. The ground truth label, the prediction, the singleton belief, the ignorance, and the highest binary confusion are noted in their captions. Each grid in the confusion matrix denotes the confusion between any two classes, while the diagonal represents the singleton belief. 

In Fig.~\ref{fig:qua_2}, the proposed method wrongly recognizes the image into {\it{dog}} class while maintaining a very high confusion towards the correct class {\it{horse}}. However, as the singleton belief is not significant for the predicted class, the proposed method could predict a combined result, \ie, both {\it{dog}} and {\it{horse}}, if setting a relatively high threshold on the sum of beliefs. Another example in Fig.~\ref{fig:qua_6} implies the ability of our method to reject making a prediction. As the image in Fig.~\ref{fig:qua_6} is contentless and low-resolution, the proposed failed to collect enough supportive evidence for any known classes, which delivers a very high ignorance value.

\subsection{Ablation Studies}
% 1. Multiple-binary head v.s. Sigmoid single-layer head. 
% 2. w. or w/o. Regularization term.
\begin{table}[t]
\scriptsize
\centering
\begin{tabular}{c|ccccc}
\hline
\multirow{3}{*}{\textbf{Method}} & \multicolumn{5}{c}{\textbf{CIFAR-10 + LSUN (crop)}} \\ \cline{2-6} 
 & \multicolumn{1}{c|}{\multirow{2}{*}{Acc.}} & \multicolumn{1}{c|}{\multirow{2}{*}{F1}} & \multicolumn{1}{c|}{\multirow{2}{*}{AUROC}} & \multicolumn{1}{c|}{AUPR} & AUPR \\
 & \multicolumn{1}{c|}{} & \multicolumn{1}{c|}{} & \multicolumn{1}{c|}{} & \multicolumn{1}{c|}{closed} & open \\ \hline\hline
Ours w/o Reg. & \multicolumn{1}{c|}{72.4} & \multicolumn{1}{c|}{84.8} & \multicolumn{1}{c|}{96.2} & \multicolumn{1}{c|}{96.9} & 95.3 \\ \hline
Ours w/ Reg. & \multicolumn{1}{c|}{\textbf{79.9}} & \multicolumn{1}{c|}{\textbf{86.9}} & \multicolumn{1}{c|}{\textbf{97.0}} & \multicolumn{1}{c|}{\textbf{97.5}} & \textbf{96.6} \\ \hline\hline
 & \multicolumn{5}{c}{\textbf{CIFAR-10 + ImageNet (crop)}} \\ \hline
Ours w/o Reg. & \multicolumn{1}{c|}{63.9} & \multicolumn{1}{c|}{82.0} & \multicolumn{1}{c|}{95.1} & \multicolumn{1}{c|}{96.1} & 93.8 \\ \hline
Ours w/ Reg. & \multicolumn{1}{c|}{\textbf{71.7}} & \multicolumn{1}{c|}{\textbf{83.9}} & \multicolumn{1}{c|}{\textbf{95.8}} & \multicolumn{1}{c|}{\textbf{96.5}} & \textbf{94.7} \\ \hline
\end{tabular}
\caption{Ablation studies about the regularization term $\mathcal{L}_{\text{reg}}$ on the task of open-set detection with two hybrid datasets.}
\label{tab:abl}
\vspace{-2mm}
\end{table}

We investigate how the regularization term $\mathcal{L}_{\text{reg}}$ influences performance. The reason we add the regularization is to avoid the first output of each plausibility function converging to the belief instead of the plausibility. To show its effectiveness, we evaluate the proposed method on the task of open-set detection in Tab.~\ref{tab:abl}. The separation could be deemed to be more sufficient if the ignorance term performs better in detecting open-set samples. Note we use the ResNet-18 as our backbone here to remain consistent with our flexible recognition experiments. For the metric of open-set detection accuracy and AUROC, closed-set samples are regarded as negative samples, while open-set samples constitute positive samples. The other two metrics, AUPR closed and AUPR open, denote the Area Under the Precision-Recall curve where closed-set or open-set images are specified as positives, respectively. The overall improvements in adding the regularization term $\mathcal{L}_{\text{reg}}$, as demonstrated in Tab.~\ref{tab:abl}, indicate its effectiveness in promoting the separation.

\subsection{Discussions and Future works}
% 1. The greedy combination is computationally expensive, which means calculating detailed confusion is not doable for large classes. But total confusion can be obtained.
% 2. Relation with conditional probabilities.
In our experiments, we find the scale of confusion and ignorance varies with different backbones and datasets. The correlation behind it could be related to the capabilities of different models and the learning difficulties of different datasets. 
% For example, on the Imagenet dataset, using ResNet-50 as the backbone reduces the overall confusion compared to ResNet-18.
We leave the investigation for our future work.

% decreases with more training classes while ignorance increases. The reason behind it could be the unexpected behavior of evidence combination under the rule of DST when dealing with a large number of sources~\cite{zadeh1986simple}.

%------------------------------------------------------------------------
\section{Conclusions}
\label{sec:conclusion}
In this paper, we introduce a novel approach to explicitly model two distinct sources of uncertainties, \ie, confusion and ignorance, under the novel task of flexible recognition. The recognition system is expected to reject samples from unknown classes and also make multiple predictions when it is uncertain about a closed-set image. Particularly, in the proposed method, the confusion is modeled as the conflicting evidence, while the ignorance represents the total lack of evidence. The hypothesis space of recognition is then divided and modeled by multiple plausibility functions. The model learns the concentration parameter of Dirichlet prior, which is being placed on the belief for singletons. A complete set of opinions could be generated through evidence combinations. Experiments on different datasets, along with challenging tasks of adversarial disturbance, flexible recognition, and open-set detection, confirm the effectiveness of the proposed method.

%%%%%%%%% REFERENCES
{\small
\bibliographystyle{ieee_fullname}
\bibliography{egbib}
}

\end{document}